\documentclass[runningheads]{llncs}
\usepackage{graphicx}
\usepackage{amsmath}
\usepackage{subcaption}
\usepackage[export]{adjustbox}[2011/08/13]
\begin{document}
\title{Combining Contrastive Learning and Knowledge Graph Embeddings to develop medical word embeddings for the Italian language}
\titlerunning{Combining contrastive learning and KGEs for medical word embeddings}

\author{Denys Amore Bondarenko\inst{1} \and
Roger Ferrod\inst{1}\and
Luigi Di Caro\inst{1}}
\authorrunning{D. Amore Bondarenko et al.}

\institute{University of Turin, Torino, Italy \\
\email{denys.amorebondarenko@unito.it \\
roger.ferrod@unito.it \\
luigi.dicaro@unito.it}
}
\maketitle              % typeset the header of the contribution
\begin{abstract}
Word embeddings play a significant role in today's Natural Language Processing tasks and applications. While pre-trained models may be directly employed and integrated into existing pipelines, they are often fine-tuned to better fit with specific languages or domains. In this paper, we attempt to improve available embeddings in the uncovered niche of the Italian medical domain through the combination of Contrastive Learning (CL) and Knowledge Graph Embedding (KGE). The main objective is to improve the accuracy of semantic similarity between medical terms, which is also used as an evaluation task. Since the Italian language lacks medical texts and controlled vocabularies, we have developed a specific solution by combining preexisting CL methods (multi-similarity loss, contextualization, dynamic sampling) and the integration of KGEs, creating a new variant of the loss. Although without having outperformed the state-of-the-art, represented by multilingual models, the obtained results are encouraging, providing a significant leap in performance compared to the starting model, while using a significantly lower amount of data.

\end{abstract}
\section{Introduction}
Texts have always represented a significant portion of all the clinical data produced every day in the world, from E.R. reports to clinical diary of patients, drugs prescriptions and administrative documents. Recent digitalization has paved the way for new applications by leveraging automatic data analysis. It is therefore necessary to develop tools capable of understanding the content of documents and their contextual nuances in order to be able to extract useful information. This is one of the main objectives of Natural Language Processing (NLP), which in recent years -- thanks to the deep-learning revolution -- has led to extraordinary results.
Many successes are due to what are known as foundational models, which are large neural networks that have been trained over a vast collection of unannotated data, capable of operating upon simple adaptation (or fine-tuning) on the most varied downstream tasks.

However, it is difficult to train a generic model suitable for every kind of text. For this reason, and starting from a pretrained model of the language of interest, a new specific embedding model is created for a given domain. This is done by continuing the training on a specific selection of texts.
Although less expensive than starting a new training from scratch, there are still many difficulties, especially when dealing with languages with limited resources, such as Italian, which lacks extensive corpora of freely accessible clinical texts. 
Due to limited resources, these models should be even more capable of operating in contexts of few annotations with regard to downstream tasks. 
In these cases, a more accurate representation of similarity is therefore necessary and turns out to be useful in many circumstances. For example, in \cite{fbk_idc9} the semantic similarity between medical terms has been exploited to reduce lexical variability by finding a common representation that can be mapped to IDC-9-CM. Starting from this work, and with the aim of improving the measure of semantic similarity, we have applied recent techniques of contrastive learning as a tool for representation learning, by approaching pairs of semantically similar or possibly equivalent terms (i.e. synonyms) and distancing dissimilar pairs.

Born in the Computer Vision field, contrastive learning is increasingly applied also to the NLP domain \cite{tutorial}, with still unexplored potential. However, the biggest difficulty lies in the efficient sampling of negative cases and the selection of positive examples, an even more difficult task in a low-resource language such as Italian. 

To compensate for the lack of synonyms listed in the Italian vocabularies of the Unified Medical Language System (UMLS), we directly exploit the Knowledge Graph Embedding (KGE) representation -- built from the UMLS semantic network -- by combining it with the word embeddings representation. In doing this, we modify the contrastive MS loss \cite{msloss} so that its parameters are tied to the similarity calculated on KGEs; we also exploit the context surrounding the terms (increasing in this way the positive cases), and a new BERT-derived model specifically fine-tuned on the Italian medical texts. To the best of our knowledge, this is the first time that MS loss, contexts and KGE have been combined in a single model. Although without having outperformed the state-of-the-art represented by multilingual models, the results obtained are encouraging and demonstrate the goodness of the developed approach, providing a significant leap in performance compared to the starting model while using a significantly lower amount of data than the state-of-the-art. However, further experiments and computational resources will be needed to extend the current model and to fully leverage the multilingual datasets.

Our main contributions are the following: 1) we trained a new word embeddings model by fine-tuning BERT on the Italian medical domain, 2) we leveraged different contrastive learning strategies to overcome the limited number of synonyms in Italian, 3) we integrated the knowledge of the UMLS semantic network by injecting its KGEs directly into the model or by modifying the contrastive loss.

\section{Related Works}
In the literature, there are many works that aim to specialize a word embedding model on a specific domain, like \cite{Lee2020BioBERTAP,Huang2019ClinicalBERTMC,10.1145/3458754}. Similar studies exist for Italian, for example \cite{Polignano2019AlBERToIB}, but not for the medical domain. To the best of our knowledge, there is no publicly available embedding model for the medical domain in Italian. There are several possible strategies to train new pretrained models, such as the possibility of training a model from scratch (like SciBERT \cite{Beltagy2019SciBERTAP}) with considerable associated costs, or to continue training on new domain-specific documents (BioBERT \cite{Lee2020BioBERTAP}); it is often necessary to extend also the vocabulary as done by \cite{souza2020bertimbau,arkhipov-etal-2019-tuning}.

In addition to word embeddings, the incorporation of the explicit knowledge represented in Knowledge Graphs (KGs) have recently been explored, injecting it into BERT and enriching in this way the model. Among the first works there is KnowBert \cite{KnowBert}, that with a mechanism of projections and re-contextualization combines word embeddings and knowledge graph embeddings calculated from Wikipedia and WordNet. A similar approach is followed by KeBioLM \cite{KeBioLM} albeit with a simplification of the architecture.  

Our work is directly inspired by SapBERT \cite{sapbert}, the first to use contrastive learning on UMLS synonyms to improve the representation of biomedical embeddings, and CODER \cite{Yuan2022CODERKI} which replaces the InfoNCE loss -- used by SapBERT -- with the MS loss and integrates the relational information of the UMLS semantic graph by adding a loss inspired by DistMult. The same authors have recently developed an extension of CODER (named CODER++ \cite{CODER++}) which introduces dynamic sampling that provides hard positive and negative pairs to the MS loss and outperforms previous results, becoming the new state-of-the-art model. While SapBERT and CODER are limited to decontextualized term, KRISSBERT \cite{KRISS} extends SapBERT by adding a context windows -- taken from PubMed -- around the terms and managing in this way their ambiguity. Furthermore, it incorporates the UMLS relationships, but limits itself to the taxonomic relationships of the ontology.
CODER is also available in a multilingual version, while a multilingual extension has recently been released also for SapBERT \cite{liu2021learning}. CODER++ and KRISSBERT, on the other hand, cannot be used directly in Italian.

\section{Proposal}
We start by developing our domain-specific medical embedding for the Italian language by extending the dataset of \cite{FerrodBCFDGMS21} with new documents, described in more detail in the \ref{sec:data} subsection. A new vocabulary was then trained on the corpus, recycling -- when possible -- the tokens and subtokens already in BERT; for example, while the word \textit{``pleuropolmonare"} (\textit{pleuropulmonary} in English) is tokenized as \textit{ple + uro + pol + mona + re} in BERT, with the new tokenizer the result is \textit{pleuro + polmonare} (with the correct lexicographic composition); other more common words are represented as single tokens (\textit{ipotensione} with the new model instead of the less informative \textit{ipote + ns + ione}). When it is possible, the new tokens, instead of being randomly initialized, are calculated as mean pooling of the original BERT subtokens, thus -- in the example -- the subword \textit{``polmonare"} is initialized as an average of [\textit{``pol",``mona",``re"}] embeddings.

By training a new vocabulary with 15k tokens, the size of the vocabulary changes from $31,102$ elements (original BERT) to $37,714$, where $55\%$ of the words are in common with the old tokenizer and the remaining new tokens are almost entirely ($94\%$) derivable from composition of previous subtokens. Henceforth, we will refer to the new model thus trained as \textit{ext-BERT} (extended BERT).

The creation of \textit{ext-BERT} proved to be essential for several linguistic tasks, as already highlighted in \cite{FerrodBCFDGMS21}, in particular for spelling correction (with a 5 percentage point increase in correction precision) and Named Entity Recognition (+6 points in F1 score). However, as shown in Section \ref{sec:results}, the similarity scores in the various proposed tasks are rather inaccurate. We therefore continued the training in a contrastive learning setting, as detailed in Section \ref{sec:model}.

\subsection{Data}
\label{sec:data}
Languages such as Italian are known to have a limited amount of publicly available data. Even more challenging is the medical domain, where specific clinical terminology is used. Medical terms alone can be confusing and difficult to understand, which would require the help of domain experts for data annotation. Fortunately, in the medical domain, the knowledge encoded in rich ontologies such as UMLS can be used to work with plain data. 

UMLS integrates many different biomedical vocabularies into one single ontology. The latest version of UMLS contains nearly $17$ million distinct concept names associated with $4,553,796$ distinct concepts. Each concept has a Concept Unique Identifier (CUI). These concepts are connected by a semantic network which consists of a set of semantic types, which provide a consistent categorization of all concepts represented in the UMLS Metathesaurus, and a set of semantic relations that exist between semantic types\cite{national2009umls}. On top of that, it incorporates intra-source relationships asserted by individual vocabularies. Furthermore, the collection of different names under the same concept creates an inter-source synonymous relationship. Although UMLS contains such a rich collection of knowledge, only $1.53\%$ is available in Italian\footnote{On average $1.22$ terms per concept in Italian vs $2.10$ terms per concept in English}. 

We leverage UMLS to build our dataset, in particular by focusing our attention on a subset of the Metathesaurus with three main Italian vocabularies: ICPCITA, MDRITA and MSHITA. Moreover, we only consider the concepts belonging to the following semantic types: Body Part, Organ, or Organ Component (BP), Body Substance (BS), Chemical (C), Medical Device (MD), Finding (F), Sign or Symptom (SS), Health Care Activity (HCA), Diagnostic Procedure (DP), Laboratory Procedure (LP), Therapeutic or Preventive Procedure (TPP), Pathologic Function (PF), Physiologic Function (PhF), and Injury or Poisoning (IP). By using UMLS 2021AB full release, we obtain $123,265$ terms and $86,610$ unique concepts. We will call this subset $\textrm{UMLS}_{ITA}$.

We construct our training dataset following KRISSBERT \cite{KRISS} method. First, we generate self-supervised mentions in our text corpus $\mathcal{T}$. The corpus we use contains approximately $36$ million words collected from different online sources available in Italian (Table \ref{tab:corpus}). Self-supervised mentions are generated by matching term surface forms from $\textrm{UMLS}_{ITA}$ in $\mathcal{T}$. We then extract contexts in a fixed $32$-token window around each mention. With this method, we manage to identify $20,447$ unique concepts and $26,432$ unique terms, resulting in a dataset that contains $2,161,918$ mention contexts. Special tags $[M_s]$ and $[M_e]$ are used to identify the beginning and the end of a mention. 

\begin{table}[h]
\begin{center}
\renewcommand{\arraystretch}{1.25}
\setlength{\tabcolsep}{10pt}
\begin{tabular}{l|lll}
\hline
Source             & Words      &      \\ \hline\hline
Wikipedia          & 9,068,684 & 25\% \\ \hline
Ministry of Health & 1,120,952  & 3\%  \\ \hline
Medical websites \& blogs            & 9,528,004 & 26\% \\ \hline
PubMed              & 2,242,367    & 6\%  \\ \hline
Medical Lectures              & 958,802   & 3\%  \\ \hline
E3C             & 7,660,558    & 21\%  \\ \hline
Medical Degree Thesis             & 5,762,792    & 16\%  \\ \hline\hline
TOTAL              & 36,342,159 &      \\ \hline
\end{tabular}
\caption{The corpus contains the collection of scientific pages of wikipedia-italian, divulgative web pages of the ministry of health, medical websites and blogs (such as Nurse24, MyPersonalTrainer, Dica33 etc.), material from university medical lectures, the E3C raw-dataset \cite{e3c} and degree thesis.}
\label{tab:corpus}
\end{center}
\end{table}

To train the Knowledge Graph Embeddings, we rely on semantic and taxonomic relationships between concepts in $\textrm{UMLS}_{ITA}$. Then we filter out rare and inverse relations, resulting in a dataset with $415,170$ triplets, $69,193$ entities and $171$ unique relationships. We then use $90\%$/$6\%$/$4\%$ training/test/validation split ratio. Data is split in such a way that the test and validation sets contain only entities and relations already seen in the training set.

\subsection{Model}
\label{sec:model}
Our first model is trained using contrastive learning, mapping similar entities closer together and different entities further apart. Term representation is retrieved from context encoding by averaging the representation vectors of tokens belonging to the entity. We adopt the multi-similarity loss (MS loss) function and modify it in a way that enables us to dynamically change the $\lambda$ parameter (i.e. the similarity margin) according to the similarity derived from Knowledge Graph Embeddings. Given the MS loss formula:
\begin{equation}
    \mathcal{L}_{MS} = \dfrac{1}{m}\sum_{i=1}^{m}\left\{\dfrac{1}{\alpha}\log[1 + \sum_{k\in\mathcal{P}_{i}}e^{-\alpha(S_{ik}-\lambda)}] + \dfrac{1}{\beta}\log[1 + \sum_{k\in\mathcal{N}_{i}}e^{\beta(S_{ik}-\lambda)}]\right\}
\end{equation}
where $\lambda$ is a fixed similarity margin. This margin heavily penalizes positive pairs with similarity $< \lambda$ and negative pairs with similarity $> \lambda$. The idea of separating positive and negative thresholds was first introduced by Liu et al.\cite{generalPairLoss}. Given the results reported by their study, we have chosen to split the threshold as follows:
\begin{equation}
    \mathcal{L}_{MS} = \dfrac{1}{m}\sum_{i=1}^{m}\left\{\dfrac{1}{\alpha}\log[1 + \sum_{k\in\mathcal{P}_{i}}e^{-\alpha(S_{ik}-\lambda_{p})}] + \dfrac{1}{\beta}\log[1 + \sum_{k\in\mathcal{N}_{i}}e^{\beta(S_{ik}-\lambda_{n})}]\right\}
\end{equation}
We will refer to this version of the loss as MS loss v2. After setting $\lambda_{p}=1$ and $\lambda_{n}=0.5$ we immediately notice improvements across all metrics. We then propose a further extension of the MS loss that exploits the similarities between KGE entities in order to dynamically chose $\lambda$. We name the following loss MS loss v3:
\begin{equation}
    \begin{split}
        \mathcal{L}_{MS} = \dfrac{1}{m}\sum_{i=1}^{m}\Bigg\{& \dfrac{1}{\alpha}\log[1 + \sum_{k\in\mathcal{P}_{i}}e^{-\alpha(S_{ik}-\lvert S_{ik}-S_{ik}^{KGE}\rvert)}] \\ 
        + & \dfrac{1}{\beta}\log[1 + \sum_{k\in\mathcal{N}_{i}}e^{\beta(S_{ik}-(1-\lvert S_{ik}-S_{ik}^{KGE}\rvert))}]\Bigg\}
    \end{split}
\end{equation}
where $S^{KGE}_{ik}$ is the similarity between concepts $i$ and $k$ in the KGE space. According to Wang and Liu\cite{wang2021understanding} the excessive pursuit to uniformity can make the contrastive loss not tolerant to semantically similar samples which may be harmful. Thus, instead of pushing all the different instances indiscriminately apart, $S^{KGE}$ helps to introduce a factor that takes into account the underlying relations between samples. The hard positive and hard negative in-batch mining is kept unchanged as in regular MS loss.

Hence, we proceed to train the model with MS loss v3. Each training batch is constructed dynamically by sampling a virtual batch from a subset $\mathcal{P}$ of our dataset $\mathcal{D}$. $\mathcal{P}$ is constructed beforehand, by selecting representative contexts of each concept. We will call these concepts prototypes. For each entity $e$ a small number of prototypes is chosen in a way that prioritizes -- where possible -- a different synonym of $e$ for each prototype. Then, for each prototype $p$ in the virtual batch, we sample $k$ positive pairs randomly, prioritizing contexts that use different synonyms than the one used in $p$. Subsequently, we sample $m$ possibly hard negative pairs, following the method introduced in \cite{CODER++}. For top-$m$ similarity search we use Faiss index, that stores embeddings of all mentions from $\mathcal{D}$ and efficiently searches for the $m$ most similar entries with respect to $p$. We update this index after each training epoch. 

In the second model (which we will call from now on ``\textit{KGE-injected}"), we use the knowledge more directly, by fusing BERT and KGE representations in the upper layers of BERT. The method is similar to the one used in KeBioLM \cite{KeBioLM}. We inject the knowledge at the layer $i$ by running the first $i$ layers of BERT and then, for each mention $m$ in the sequence, we apply the mean pool to the tokens of $m$, obtaining the BERT mention representation $\textbf{h}_m$. Then a linear projection is used to map each mention to the KGE space $\textbf{h}_m^{proj}=\textbf{W}_m^{proj}\textbf{h}_m+\textbf{b}^{proj}$. We then use an entity linker, which selects $n$ candidate entities closest to $h_m^{proj}$. The similarities of the $n$ candidates are normalized through a softmax function. This gives us the normalized similarity scores $\textbf{a}$ that are used to compute the combined entity representation:
\begin{equation}
    e_m = \sum_{j=1}^{n}a_j\cdot \textbf{e}_j
\end{equation}
where $\textbf{e}_j$ is the knowledge graph embedding of the entity $j$.

Unlike KeBioLM, we keep entity embeddings $\textbf{e}$ fixed throughout the training. After obtaining the entity representation, we project it back to the BERT embedding space, where it is added to every token of the mention. The resulting embeddings are then normalized and forwarded to the following levels of BERT as usual. 

To link the mentions encoded by BERT to the KGE entities, we define an entity linking loss as cross-entropy between self-supervised entity labels and similarities obtained from the linker in KGE space:
\begin{equation}
    \mathcal{L}_{EL}=\sum-\log\dfrac{\exp(h_m^{proj}\cdot\textbf{e})}{\sum_{\textbf{e}_j\in \mathcal{E}} \exp(h_m^{proj}\cdot\textbf{e}_j)}
\end{equation}
where $\mathcal{E}$ is the KGE entity set.

Furthermore, we add the masking language modeling task to prevent the catastrophic forgetting phenomenon \cite{McCloskey1989CatastrophicII,dAutume2019EpisodicMI}. As done in \cite{KnowBert}, we mask the whole entity if any of the $15\%$ masked tokens happens to belong to a mention. Ultimately, we jointly minimize the following loss:
\begin{equation}
    \mathcal{L} = \mathcal{L}_{MLM} + \mathcal{L}_{EL}
\end{equation}

The two previously described models use knowledge in different ways, improving particular aspects of the representation and returning different results depending on the task. Therefore, we decided to combine both in a single model, using the KGE injection training as pre-training phase and the contrastive learning with MS loss v3 as fine-tuning process. We call the latter model \textit{``pipelined"}.

\section{Results}
\label{sec:results}
All the training is performed on one \textit{NVIDIA T4} GPU, which has $16$ GB of memory. For this reason, we could not experiment with larger batches, like those used for training CODER and CODER++, which were trained on $8$ \textit{NVIDIA A100} $40$GB GPUs. 

We evaluated our models on three similarity-oriented metrics: MSCM score, clustering pair and semantic relatedness. MSCM is a similarity score based on the UMLS taxonomy, developed by \cite{mcsm} and used in CODER. It is defined as:
\begin{equation}
MSCM(V,T,k) = \frac{1}{V(T)}\sum_{v \in V(T)}{\sum_{i=1}^{k}{\frac{1_T(v(i))}{log_2(i+1)}}}
\end{equation}
where $V$ is a set of concepts, $T$ the semantic type according to UMLS, $k$ the parameterized number of neighborhood, $V(T)$ a subset of concepts with type $T$, $v(i)$ the $i^{th}$ closest neighbor of concept $v$ and $1_T$ is an indicator function which is $1$ if $v$ is of type $T$, $0$ otherwise. Given this formulation and the default settings ($k=40$ as used in CODER) the score ranges from $0$ to $11.09$.
Given its importance in low-resource language, where pre-trained tools for entity recognition and linking are lacking, we have also included the clustering pair task. Already experimented in \cite{fbk_idc9} to unify lexically different but semantically equivalent terms, the task is defined more formally by CODER++, where two terms are considered synonyms if their cosine similarity is higher than a given threshold ($\theta$) meanwhile true synonyms are taken from UMLS.
For semantic relatedness, since there are no datasets of this kind for Italian and given their development costs (which would require the intervention of several domain experts), we rely on two English datasets, manually translating the entities involved. MayoSRS and UMNSRS were introduced by \cite{mayosrs} and \cite{umnsrs} with a manual annotation of a relatedness score for 101 and 587 medical term pairs, respectively. The values vary from 1 to 10 for MayoSRS and 0-1600 for UMNSRS. Due to the lack of an appropriate translation for some terms, the number of pairs for the UMNSRS dataset is reduced to 536 tuples.

A first comparison of the state-of-the-art shows that SapBERT, in the multilingual version, actually outperforms CODER, despite the fact that the latter was in advantage in the paper that introduced it.
As regards the training of KGEs, Table \ref{tab:kge} shows the results on the different models evaluated on the link prediction and similarity tasks. We have chosen \textit{ComplEx} as reference model, thanks to the good results obtained on the similarity datasets and a representation still comparable with the other models.

The use of \textit{ComplEx} in MS loss v3 proved to be of substantial benefit, as shown in Table \ref{tab:losses} where we compare the performances obtained with the different variants of the loss. \textit{ComplEx} also proved to be the superior embedding for the KGE-injected model, obtaining a moderate improvement relative to the baseline (\textit{ext-BERT}) albeit in a more contained way if compared to the contrastive learning training. 

\begin{table}[h]
\begin{center}
\begin{tabular}{llllll|lll}
\hline
model                        & hits@1        & hits@3        & hits@10       & MR              & MRR           & MCSM          & MayoSRS       & UMNSRS        \\ \hline \hline
\multicolumn{1}{l|}{TransE}  & 0.07          & 0.21          & 0.38          & 1619         & \textbf{0.17} & 9.76          & 0.45          & \textbf{0.49} \\
\multicolumn{1}{l|}{ComplEx} & 0.07          & 0.19          & 0.34          & 1918         & 0.16          & \textbf{9.96} & \textbf{0.55} & 0.45          \\
\multicolumn{1}{l|}{RotatE}  & \textbf{0.14} & \textbf{0.25} & \textbf{0.42} & \textbf{3382} & \textbf{0.17} & 9.42          & 0.52          & 0.40          \\
\multicolumn{1}{l|}{SimplE}  & 0.09          & 0.17          & 0.30          & 2608         & 0.16          & 9.68          & 0.47          & 0.41          \\ \hline
\end{tabular}
\caption{Evaluations of Knowledge Graph Embeddings over link prediction task (hits@k, Mean Rank, Mean Reciprocal Rank) and similarity tasks (MSCM score and Spearman coefficient over relatedness scores).}
\label{tab:kge}
\end{center}
\end{table}

\begin{table}[h]
\begin{center}
\begin{tabular}{l|llll}
\hline
model      & \multicolumn{1}{c}{\begin{tabular}[c]{@{}c@{}}MCSM\\ (avg)\end{tabular}} & \multicolumn{1}{c}{\begin{tabular}[c]{@{}c@{}}Clustering \\    (F1)\end{tabular}} & MayoSRS & UMNSRS \\ \hline \hline
MS loss v1 & 5.29                                                                     & 14.98                                                                              & \textbf{0.36}    & 0.26   \\
MS loss v2 & 5.36                                                                     & 16.91                                                                             & 0.30    & 0.27   \\
MS loss v3 & \textbf{5.57}                                                                     & \textbf{17.44}                                                                             & 0.30    & \textbf{0.36}   \\ \hline
\end{tabular}
\caption{Comparison between training done with the original MS loss (v1), MS loss with separate margins (v2) and our proposal (v3).}
\label{tab:losses}
\end{center}
\end{table}

Finally, we combine the two models, exploiting the contrastive learning (with MS loss v3) and the previously trained \textit{KGE-injected} model. The results thus obtained, shown in Tables \ref{tab:mcsm} and \ref{tab:tables}, are better than the previous models taken individually.
To obtain the final model, we first train \textit{ext-BERT} with the KGE injection approach for $40$k training steps. Each batch contains $6$ training sequences, where each sequence contains at least $1$ and at most $119$ mentions. KGEs are then injected at the 8\textsuperscript{th} BERT layer. The remaining hyperparameters are the following: $4$k warm-up steps, weight decay $0.01$, and learning rate $1e-5$. 
The resulting model is then trained with MS loss v3 for $50$k training steps (~4 epochs). For each training step, we sample $4$ prototypes $p$ from $\mathcal{P}$. We set the number of positives as $k=20$, and the number of possible negatives as $m=30$. 
With regard to the (possible) negatives mining, we update the Faiss index at every epoch. We also experiment with different MS loss parameters, but without seeing any improvement and thus leaving the original $\alpha=2$, $\beta=50$, $\epsilon=0.1$. Other parameters are: learning rate $2e-5$, weight decay $0.01$, max gradient norm $1$ and $20$k warm-up steps. During the training, we use gradient accumulation for $8$ steps, while the number of contexts per term is limited to $4$ for computational reasons. 

The chosen parameters represent a compromise between the performances obtained on the various tasks. In fact, we have noticed a different behavior of the model depending on the task on which it is evaluated. In particular, the human annotated semantic relatedness seems to be in contrast with the metrics defined automatically from UMLS; each improvement of human metrics corresponds to a worsening of UMLS-based metrics and vice versa. Moreover, while the clustering pair task seems to benefit particularly from the increase in the number of epochs \footnote{with 16 epochs the F1 score stands at 25.62, closer to 33.92 of the SOTA model than 19.37 of the 4 epochs model}, the prolonged training has a slightly negative effect on semantic relatedness and strongly penalizes MSCM score. The choice of the pooling strategy is also not optimal for all tasks. By replacing the mean pooling with the CLS tag, both during training and validation, we obtain higher than state-of-the-art scores on the MayoSRS and UMNSRS datasets\footnote{respectively 0.49 and 0.50 vs 0.44 and 0.48 of the SOTA model}, however this choice is ineffective for MSCM (-0.61 percentage points on average) and clustering pair (-10.39 points). Finally, we observed that the Masked Language Modeling training is counterproductive with respect to similarity measures. In fact, by comparing \textit{ext-BERT} with the basic version of BERT, it is already evident that performances dropped over all datasets, despite having obtained significant gains in linguistic tasks. At the same time, the new representation obtained with contrastive learning does not seem to bring any benefit on linguistic tasks, as happened with CODER and SapBERT. This phenomenon needs to be further investigated in order to find the right balance.

\begin{table}[]
\begin{center}
\begin{adjustbox}{width=1.2\textwidth,center=\textwidth}
\begin{tabular}{l|ccccccccccccc|ll}
\hline
Model      & BP & BS & C.                             & MD                               & F                              & SS & HCA                             & DP & LP &  TPP & PF & PhF & IPs & AVG  &       \\ \hline \hline
mBERT      & 3,19                                 & 0,76                                 & 8,98                                  & 2,36                                 & 6,65                                 & 3,72                                 & 4,02                                 & 3,43                                 & 3,04                         & 6,17                                 & 9,07                                & 2,85                                 & 6,27                                   & 4.65 & --    \\
BERT       & 2,97                                 & 0,78                                 & 8,72                                  & 2,57                                 & 6,3                                  & 3,27                                 & 3,84                                 & 3,81                                 & 2,86                         & 6                                    & 9,07                                & 2,66                                 & 6                                      & 4.53 & --    \\ \hline
SapBERT    & {\textbf{6,06}} & {\textbf{1,79}} & {\textbf{10,19}} & {\textbf{4,38}} & { \textbf{7,54}} & { \textbf{4,82}} & { \textbf{5,48}} & { \textbf{6,69}} & { 4,39}  & { \textbf{7,92}} & { 9,52}         & { \textbf{4,46}} & { 6,98}            & \textbf{6.17} & +33\% \\
CODER      & 4,1                                  & 1,22                                 & 9,63                                  & 2,99                                 & 6,57                                 & 3,94                                 & 4,95                                 & 4,01                                 & 3,2                          & 6,04                                 & 9,08                                & 3,29                                 & 5,87                                   & 4.99 & +7\%  \\
ext-BERT   & 2,84                                 & 0,93                                 & 8,71                                  & 2,37                                 & 6,72                                 & 3,28                                 & 3,06                                 & 3,48                                 & 3,13                         & 6,04                                 & 9,05                                & 2,37                                 & 6,21                                   & 4.48 & -1\%  \\ \hline
MS loss v3         & 4,38                                 & 1,58                                 & 9,77                                  & 3,77                                 & 7,05                                 & 4,37                                 & 4,41                                 & 5,37                                 & 4,25                         & 7,49                                 & 9,57                                & 3,37                                 & 6,98                                   & 5.57 & +24\% \\
KGE-injected        & { 3}             & { 0,9}           & { 8,81}           & { 2,64}          & { 7,04}          & { 3,65}          & { 3,96}          & { 3,91}          & { 3,01}  & { 6,36}          & { 9,09}         & { 2,93}          & { 6,81}            & 4.78 & +7\%  \\
pipelined & 4,84                                 & 1,58                                 & 9,67                                  & 3,72                                 & 7,31                                 & 4,6                                  & 5,02                                 & 5,83                                 & \textbf{4,66}                & 7,81                                 & \textbf{9,59}                       & 3,69                                 & { \textbf{7,36}}   & 5.82 & +30\% \\ \hline
\end{tabular}
\end{adjustbox}
\caption{Evaluation of baselines, state-of-the-art and our models with the MSCM similarity score.}
\label{tab:mcsm}
\end{center}
\end{table}

\begin{table}
\begin{subtable}[c]{0.5\textwidth}
\centering
\begin{tabular}{l|lllll}
\hline
Model      & \multicolumn{1}{c}{$\theta$}       & \multicolumn{1}{c}{A}      & \multicolumn{1}{c}{F1}       & \multicolumn{1}{c}{P}        & \multicolumn{1}{c}{R}        \\ \hline \hline
mBERT      & \multicolumn{1}{r}{0,93}    & \multicolumn{1}{r}{100}    & \multicolumn{1}{r}{6,4}      & \multicolumn{1}{r}{4,8}      & \multicolumn{1}{r}{9,59}     \\
BERT       & \multicolumn{1}{r}{0,94}    & \multicolumn{1}{r}{100}    & \multicolumn{1}{r}{7,91}     & \multicolumn{1}{r}{5,52}     & \multicolumn{1}{r}{13,94}    \\ \hline
SapBERT    & {0,88} & {100} & {\textbf{33,92}} & {\textbf{35,83}} & {32,21} \\
CODER      & 0,87                        & 100                        & 32,24                        & 31,86                        & {\textbf{32,64}} \\
ext-BERT   & 0,94                        & \multicolumn{1}{r}{100}    & \multicolumn{1}{r}{5,00}        & \multicolumn{1}{r}{3,86}     & \multicolumn{1}{r}{7,22}     \\ \hline
MS loss v3         & 0,88                        & 100                        & 17,44                        & 13,75                        & 23,83                        \\
KGE-injected        & {0,90}  & {100} & {5,86}  & {4,09}  & {10,32} \\
pipelined & 0,89                        & 100                        & 19,37                        & 15,55                        & 25,68                        \\ \hline
\end{tabular}
\subcaption{Clustering pair}
\label{tab:clustering}
\end{subtable}
\begin{subtable}[c]{0.5\textwidth}
\centering
\begin{tabular}{l|cc}
\hline
Model      & \multicolumn{1}{c}{MayoSRS}          & \multicolumn{1}{c}{UMNSRS}                      \\ \hline \hline
mBERT      & 0,00             & 0,14                                            \\
BERT       & 0,12             & 0,19                                            \\ \hline
SapBERT    & {0,37}          & {0,33}                     \\
CODER      & { \textbf{0,44}} & { \textbf{0,48}}            \\
ext-BERT   & -0,07                                & 0,24                                            \\ \hline
Ms loss v3         & 0,30             & 0,36                                            \\
KGE-injected        & {0,31}          & 0,32 \\
pipelined & 0,32                                 & 0,41                      \\ \hline
\end{tabular}
\subcaption{Relatedness score}
\label{tab:similarity}
\end{subtable}
\caption{a) Results of the clustering pair task (i.e. automatic detection of synonyms); the value of the threshold $\theta$ is optimized over the F1 score.
b) Spearman coefficient over two semantic relatedness datasets: MayoSRS (with range 1-10) and UMNSRS (0-1600).
}
\label{tab:tables}
\end{table}

\section{Conclusion and Future works}
In this paper, we attempt to reorganize the representation of the embedding space in order to improve the measure of similarity between medical word embeddings in the Italian language. 
With regard to metric learning and representation learning, contrastive learning is currently the most suitable and widely adopted method, which aims to bring similar terms together and distancing dissimilar cases. We used the MS loss (already experimented in CODER) and its hard negatives sampling mechanism, the contextualization of the terms (introduced in KRISSBERT) and the dynamic sampling of CODER++ (i.e. the mining of top-k similarities as possibly hard negative samples), combining for the first time all these elements together. 

However, unlike other works, models trained on the Italian language have to overcome more substantial challenges. The use of contexts has proved to be essential, not only to capture different nuances of the same term, but in general to expand the number of positives, which would have been too few for a successful training if we had limited ourselves to synonyms. To this are added the computational difficulties, since contrastive learning is a notoriously onerous task. To overcome these difficulties, and leverage as much as possible the information available, we have exploited to our advantage the information contained in the KGEs, either by injecting them directly into the word embeddings model, or by adapting the MS loss in order to take into account also the similarity calculated on them. The latter contribution represents the major novelty of this work, as
such an initiative had never been proposed before; it has allowed, in our case, a considerable increase in performances, approaching the SOTA models despite having much fewer data and computer power.

The fact of not having outperformed the state-of-the-art of multilingual models suggests to us that there can be an advantage in moving to a multilingual environment. Probably because, in a multilingual setting, other languages can be leveraged in the absence of synonyms and terms in the less represented language. An extension of this type will certainly require more computing resources and more contexts to sample. Therefore, we intend to experiment with this path by selecting a limited set of languages in the near future. Eventually also the sampling of negatives could be avoided by changing paradigm and abandoning contrastive learning; for example, works based on the redundancy-reduction principle (e.g. \cite{Zbontar2021BarlowTS}) have recently shown results comparable to traditional contrastive learning methods, by modifying the type of loss and renouncing the sampling of negative examples. However, to the best of our knowledge, there are still no such works in the NLP field. We keep this direction as future work to be tested. 

\bibliographystyle{splncs04}
\bibliography{biblio}

\begin{thebibliography}{10}
\providecommand{\url}[1]{\texttt{#1}}
\providecommand{\urlprefix}{URL }
\providecommand{\doi}[1]{https://doi.org/#1}

\bibitem{arkhipov-etal-2019-tuning}
Arkhipov, M., Trofimova, M., Kuratov, Y., Sorokin, A.: Tuning multilingual
  transformers for language-specific named entity recognition. In: Proceedings
  of the 7th Workshop on Balto-Slavic Natural Language Processing. pp. 89--93.
  Association for Computational Linguistics, Florence, Italy (Aug 2019).
  \doi{10.18653/v1/W19-3712}, \url{https://www.aclweb.org/anthology/W19-3712}

\bibitem{Beltagy2019SciBERTAP}
Beltagy, I., Lo, K., Cohan, A.: Scibert: A pretrained language model for
  scientific text. In: EMNLP (2019)

\bibitem{mcsm}
Choi, Y., Chiu, C.Y.I., Sontag, D.A.: Learning low-dimensional representations
  of medical concepts. AMIA Summits on Translational Science Proceedings
  \textbf{2016},  41 -- 50 (2016)

\bibitem{FerrodBCFDGMS21}
Ferrod, R., Brunetti, E., Caro, L.D., Francescomarino, C.D., Dragoni, M.,
  Ghidini, C., Marinello, R., Sulis, E.: A support for understanding medical
  notes: Correcting spelling errors in italian clinical records. In:
  SMARTERCARE@AI*IA. pp. 19--28 (2021),
  \url{http://ceur-ws.org/Vol-3060/paper-3.pdf}

\bibitem{10.1145/3458754}
Gu, Y., Tinn, R., Cheng, H., Lucas, M., Usuyama, N., Liu, X., Naumann, T., Gao,
  J., Poon, H.: Domain-specific language model pretraining for biomedical
  natural language processing. ACM Trans. Comput. Healthcare  \textbf{3}(1)
  (oct 2021). \doi{10.1145/3458754}, \url{https://doi.org/10.1145/3458754}

\bibitem{Huang2019ClinicalBERTMC}
Huang, K., Altosaar, J., Ranganath, R.: Clinicalbert: Modeling clinical notes
  and predicting hospital readmission. ArXiv  \textbf{abs/1904.05342} (2019)

\bibitem{Lee2020BioBERTAP}
Lee, J., Yoon, W., Kim, S., Kim, D., Kim, S., So, C.H., Kang, J.: Biobert: a
  pre-trained biomedical language representation model for biomedical text
  mining. Bioinformatics  \textbf{36},  1234 -- 1240 (2020)

\bibitem{sapbert}
Liu, F., Shareghi, E., Meng, Z., Basaldella, M., Collier, N.: Self-alignment
  pretraining for biomedical entity representations. In: NAACL (2021)

\bibitem{liu2021learning}
Liu, F., Vuli{\'c}, I., Korhonen, A., Collier, N.: Learning domain-specialised
  representations for cross-lingual biomedical entity linking. In: Proceedings
  of ACL-IJCNLP 2021 (Aug 2021)

\bibitem{generalPairLoss}
Liu, H., Cheng, J., Wang, W., Su, Y.: The general pair-based weighting loss for
  deep metric learning. arXiv preprint arXiv:1905.12837  (2019)

\bibitem{e3c}
Magnini, B., Altuna, B., Lavelli, A., Speranza, M., Zanoli, R.: The e3c
  project: European clinical case corpus. In: SEPLN (2021)

\bibitem{dAutume2019EpisodicMI}
de~Masson~d'Autume, C., Ruder, S., Kong, L., Yogatama, D.: Episodic memory in
  lifelong language learning. ArXiv  \textbf{abs/1906.01076} (2019)

\bibitem{McCloskey1989CatastrophicII}
McCloskey, M., Cohen, N.J.: Catastrophic interference in connectionist
  networks: The sequential learning problem. Psychology of Learning and
  Motivation  \textbf{24},  109--165 (1989)

\bibitem{national2009umls}
{National Library of Medicine (US)}: UMLS® Reference Manual [Internet]. NCBI
  (2009), \url{https://www.ncbi.nlm.nih.gov/books/NBK9676/}

\bibitem{umnsrs}
Pakhomov, S., McInnes, B., Adam, T., Liu, Y., Pedersen, T., Melton, G.:
  Semantic similarity and relatedness between clinical terms: An experimental
  study. AMIA ... Annual Symposium proceedings / AMIA Symposium. AMIA Symposium
   \textbf{2010},  572--576 (Nov 2010)

\bibitem{mayosrs}
Pakhomov, S.V.S., Pedersen, T., McInnes, B.T., Melton, G.B., Ruggieri, A.P.,
  Chute, C.G.: Towards a framework for developing semantic relatedness
  reference standards. Journal of biomedical informatics  \textbf{44 2},
  251--65 (2011)

\bibitem{KnowBert}
Peters, M.E., Neumann, M., RobertL.Logan, I., Schwartz, R., Joshi, V., Singh,
  S., Smith, N.A.: Knowledge enhanced contextual word representations. In:
  EMNLP (2019)

\bibitem{Polignano2019AlBERToIB}
Polignano, M., Basile, P., Degemmis, M., Semeraro, G., Basile, V.: Alberto:
  Italian bert language understanding model for nlp challenging tasks based on
  tweets. In: CLiC-it (2019)

\bibitem{fbk_idc9}
Ronzani, M., Ferrod, R., Di~Francescomarino, C., Sulis, E., Aringhieri, R.,
  Boella, G., Brunetti, E., Di~Caro, L., Dragoni, M., Ghidini, C., Marinello,
  R.: Unstructured data in predictive process monitoring: Lexicographic and
  semantic mapping to icd-9-cm codes for the home hospitalization service. In:
  AIxIA 2021 – Advances in Artificial Intelligence: 20th International
  Conference of the Italian Association for Artificial Intelligence, Virtual
  Event, December 1–3, 2021, Revised Selected Papers. p. 700–715.
  Springer-Verlag, Berlin, Heidelberg (2021)

\bibitem{souza2020bertimbau}
Souza, F., Nogueira, R., Lotufo, R.: Bertimbau: Pretrained bert models for
  brazilian portuguese. In: Cerri, R., Prati, R.C. (eds.) Intelligent Systems.
  pp. 403--417. Springer International Publishing, Cham (2020)

\bibitem{wang2021understanding}
Wang, F., Liu, H.: Understanding the behaviour of contrastive loss. In:
  Proceedings of the IEEE/CVF conference on computer vision and pattern
  recognition. pp. 2495--2504 (2021)

\bibitem{msloss}
Wang, X., Han, X., Huang, W., Dong, D., Scott, M.R.: Multi-similarity loss with
  general pair weighting for deep metric learning. 2019 IEEE/CVF Conference on
  Computer Vision and Pattern Recognition (CVPR) pp. 5017--5025 (2019)

\bibitem{KeBioLM}
Yuan, Z., Liu, Y., Tan, C., Huang, S., Huang, F.: Improving biomedical
  pretrained language models with knowledge. In: BIONLP (2021)

\bibitem{Yuan2022CODERKI}
Yuan, Z., Zhao, Z., Yu, S.: Coder: Knowledge infused cross-lingual medical term
  embedding for term normalization. Journal of biomedical informatics p. 103983
  (2022)

\bibitem{Zbontar2021BarlowTS}
Zbontar, J., Jing, L., Misra, I., LeCun, Y., Deny, S.: Barlow twins:
  Self-supervised learning via redundancy reduction. In: ICML (2021)

\bibitem{CODER++}
Zeng, S., Yuan, Z., Yu, S.: Automatic biomedical term clustering by learning
  fine-grained term representations. In: BIONLP (2022)

\bibitem{tutorial}
Zhang, R., Ji, Y., Zhang, Y., Passonneau, R.J.: Contrastive data and learning
  for natural language processing. In: Proceedings of the 2022 Conference of
  the North American Chapter of the Association for Computational Linguistics:
  Human Language Technologies: Tutorial Abstracts. pp. 39--47. Association for
  Computational Linguistics, Seattle, United States (Jul 2022).
  \doi{10.18653/v1/2022.naacl-tutorials.6},
  \url{https://aclanthology.org/2022.naacl-tutorials.6}

\bibitem{KRISS}
Zhang, S., Cheng, H., Vashishth, S., Wong, C., Xiao, J., Liu, X., Naumann, T.,
  Gao, J., Poon, H.: Knowledge-rich self-supervised entity linking. ArXiv
  \textbf{abs/2112.07887} (2021)

\end{thebibliography}

\end{document}